\documentclass[journal]{IEEEtran}
\IEEEoverridecommandlockouts
\usepackage{cite}
\usepackage{amsmath,amssymb,amsfonts}
\usepackage{algorithmic}
\usepackage{graphicx}
\usepackage{textcomp}
\usepackage{xcolor}
\usepackage{url}
\usepackage[caption=false,font=normalsize,labelfont=sf,textfont=sf]{subfig}
\usepackage{stfloats}
\usepackage{verbatim}
\usepackage[export]{adjustbox}
\usepackage{graphbox}
\usepackage{makecell}
\usepackage{mathabx}
\usepackage{hyperref}

\graphicspath{{./figs}}
\DeclareGraphicsExtensions{.pdf,.jpg,.png}

\def\BibTeX{{\rm B\kern-.05em{\sc i\kern-.025em b}\kern-.08em
    T\kern-.1667em\lower.7ex\hbox{E}\kern-.125emX}}
\begin{document}

\title{Grasping, Part Identification, and Pose Refinement in One Shot with a Tactile Gripper}

\author{Joyce Xin-Yan Lim and Quang-Cuong Pham
\thanks{J.X.Y. Lim and Q.C. Pham are with the HP-NTU Digital Manufacturing Corporate Lab and the School of Mechanical and Aerospace Engineering, Nanyang Technological University, Singapore.}}


\maketitle

\begin{abstract}

The rise in additive manufacturing comes with unique opportunities and challenges. Rapid changes to part design and massive part customization distinctive to 3D-Print (3DP) can be easily achieved. Customized parts that are unique, yet exhibit similar features such as dental moulds, shoe insoles, or engine vanes could be industrially manufactured with 3DP. However, the opportunity for massive part customization comes with unique challenges for the existing production paradigm of robotics applications, as the current robotics paradigm for part identification and pose refinement is repetitive, where data-driven and object-dependent approaches are often used. Thus, a bottleneck exists in robotics applications for 3DP parts where massive customization is involved, as it is difficult for feature-based deep learning approaches to distinguish between similar parts such as shoe insoles belonging to different people. As such, we propose a method that augments patterns on 3DP parts so that grasping, part identification, and pose refinement can be executed in one shot with a tactile gripper. We also experimentally evaluate our approach from three perspectives, including real insertion tasks that mimic robotic sorting and packing, and achieved excellent classification results, a high insertion success rate of 95\%, and a sub-millimeter pose refinement accuracy.

\end{abstract}

\begin{IEEEkeywords}
additive manufacturing, tactile pose refinement, object classification 
\end{IEEEkeywords}

\section{Introduction}

\IEEEPARstart{A}{dditive} manufacturing has revolutionized the design and manufacturing of parts. Rapid changes to part design and massive part customization distinctive to 3DP can be easily achieved. Customized parts that are unique, yet exhibit similar features such as dental moulds, shoe insoles, or even engine vanes could be industrially manufactured with 3DP. However, a major drawback to 3DP in manufacturing lines arises from the need to conduct manual post-processing, such as cleaning of residue powder, painting, sorting, and packing. Thus, it is desirable to introduce robotics and automation to achieve end-to-end 3DP post-post processing, due to bottlenecks from the use of manual labor in post-processing.

A key aspect of robotics application is robot perception, where information on the environment is obtained for the robot to plan and execute motions, such as grasping and manipulation. Object information can be obtained by vision cameras to conduct object identification and pose estimation, which is usually executed with feature extraction methods, either by classical methods~\cite{lowe2004sift, rublee2011orb}, or deep learning methods~\cite{dong2019ppr, michel2017global,li2018deepim}.

Apart from using vision cameras, there are also studies on tactile perception for pose estimation and object classification that were propelled by the introduction of vision-based tactile sensors such as GelSight~\cite{yuan2017gelsight} and Digit~\cite{lambeta2020digit}. These sensors provide information on the contact geometry of the object and can be more informative than traditional tactile sensors that detect force or pressure distributions~\cite{dahiya2009tactile, yousef2011tactile} because local force and shear can be inferred from the high-resolution tactile image of the contact geometry. Recent works include visual servoing~\cite{chaudhury2022using}, filter-based methods for sparse point cloud registration~\cite{murali2021active} or learning-based methods~\cite{villalonga2021tactile, bauza2019localization, bauza2022tac2pose, lin2019learning}. Other works on tactile exploration include object shape estimation~\cite{matsubara2017active} and shape estimation for grasp planning~\cite{de2021simultaneous}.

\begin{figure}[t!]
    \centering
	\includegraphics[width=0.37\textwidth]{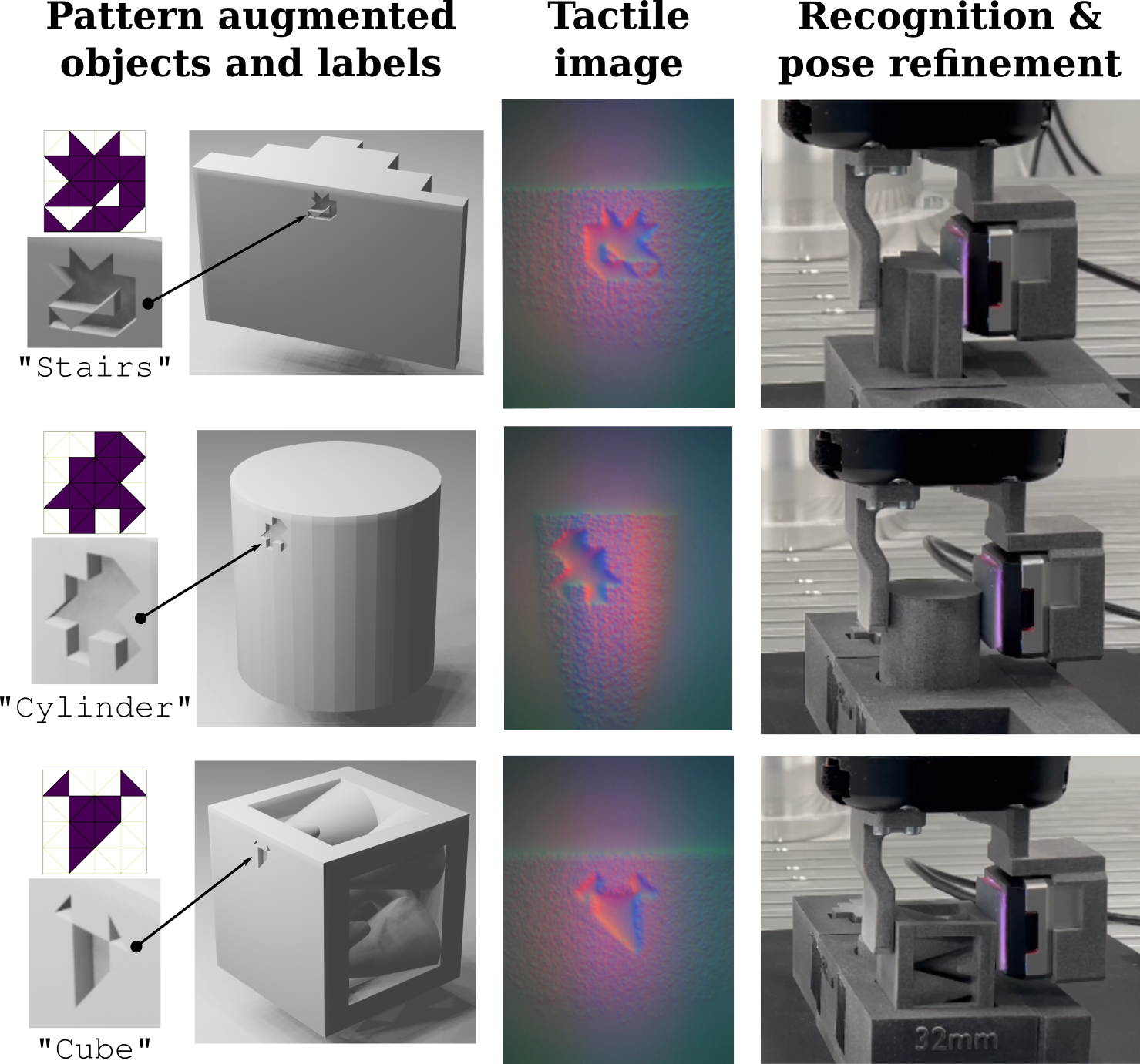}
        \caption{Pattern augmentation on 3DP parts for object recognition and high accuracy pose refinement to conduct insertion tasks. A video demonstration is available at \protect\url{https://youtu.be/3e6gvkZUk8c}}
        \label{fig:overview}
\end{figure}

However, in addition to the bottleneck created due to manual post-processing, the opportunity of massive customization comes with unique challenges for the existing production paradigm of robotics applications. Current paradigms for part identification and pose refinement are repetitive, where parts with unique features could be identified using feature-based deep learning methods, or by labeling parts, grasping them, and presenting them to a camera to identify the labels. With massive part customization, feature-based deep learning approaches have difficulties in differentiating similar parts such as shoe insoles that belong to different people, leading to limitations in end-to-end 3DP post-processing automation.

Thus, we aim to support end-to-end 3DP post-processing automation by exploring the use of pattern augmentation on 3DP objects to execute grasping, part identification, and pose refinement in one shot with a tactile gripper that can achieve fast, excellent part identification, and high object insertion success rate of 95\% with sub-millimeter accuracy (Fig.~\ref{fig:overview}), which is the first to the best of our knowledge. Our approach leverages the advantage of 3DP since the objects are to be manufactured by 3DP, and pattern augmentation allows unique patterns to correspond to objects thus enabling differentiation between similar objects. Additionally, upon extraction of the tactile imprint, part identification and pose refinement were achieved in only 0.4s. A unique advantage of our method is that grasping, part identification, and pose refinement are conducted simultaneously, unlike the common sequential process where a robot has to bring a grasped part to a camera.

The rest of the paper is as follows: Section~\ref{sec:related_work} reviews related works, Section~\ref{sec:method} introduces our method, and Section~\ref{sec:experiments} experimentally evaluates our approach from three perspectives.

\section{Related work} \label{sec:related_work}

\begin{figure*}[htp] 
    \centering
    \includegraphics[width=0.75\linewidth]{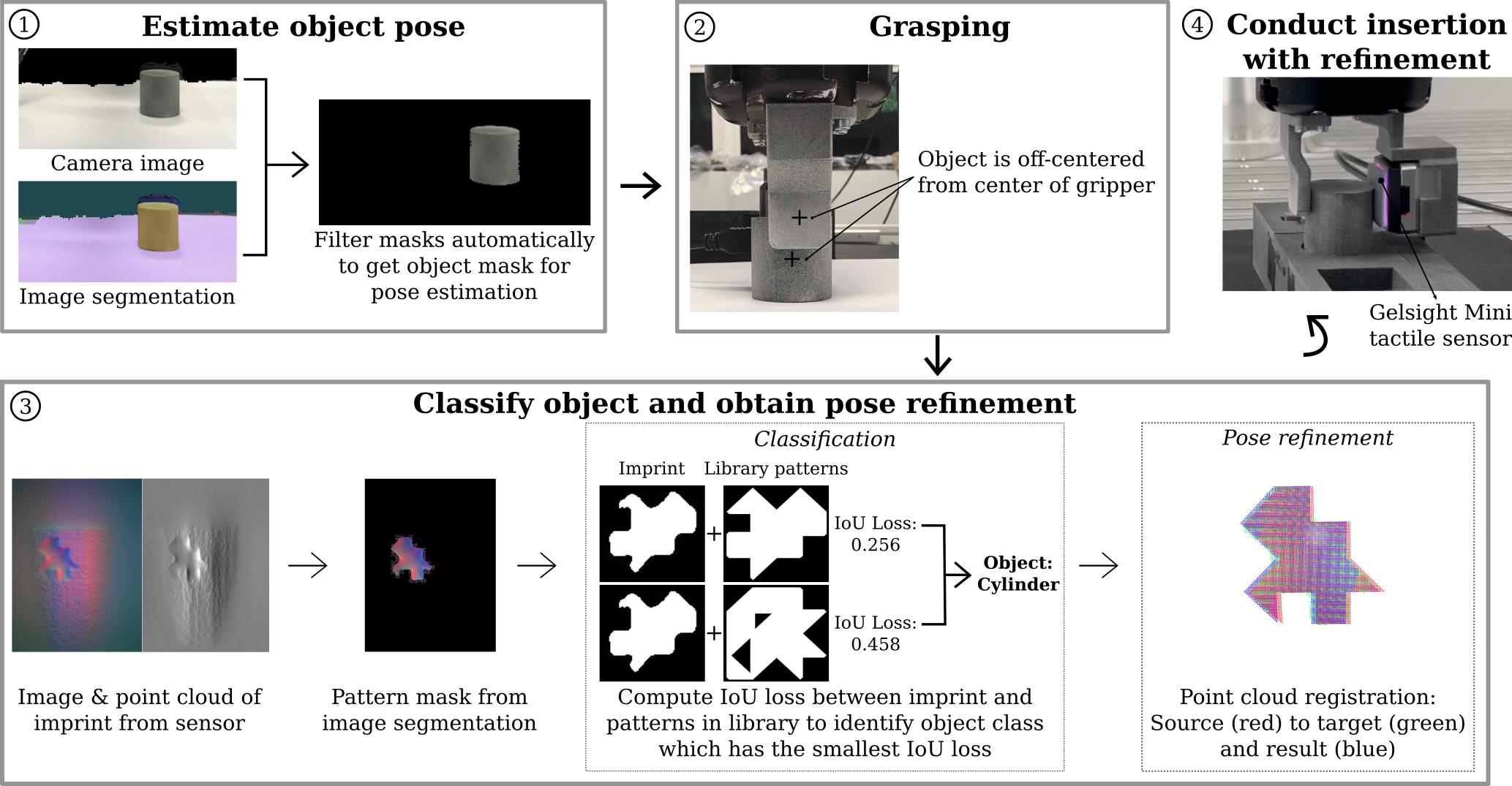} 
    \caption{Graphical pipeline for object classification and pose refinement for pattern augmented 3DP objects.} 
  \label{fig:pipeline}
\end{figure*}

Robot perception is a wide area of research. This section focuses on reviewing previous works on feature-based approaches to conduct localization of 3DP objects and works on tactile perception, which are more relevant.

Due to the customization of parts in 3DP, generalized feature-based approaches may be more suitable to reduce object dependency on applications. Feature-based approaches have been widely used for localization by the extraction of keypoints, with classical~\cite{lowe2004sift, rublee2011orb} or deep learning methods~\cite{lfnet, superpoint}. These features could be used in visual-servoing, where synthetic target images are used to steer the manipulator to the desired position for grasping 3DP parts~\cite{dfbvs}. Another work proposed a point-pair feature descriptor to estimate the pose of industrial objects in a bin, and the minimum pose error is 10\% of the maximum dimension~\cite{s18082719}. However, the main weaknesses of these approaches are that high accuracy and consistency are quite difficult to achieve. In addition, parts that are unique yet exhibit similar features cannot be differentiated.

Vision-based tactile sensors have been widely incorporated in robotics research for object localization, pose estimation, and object shape exploration. Recent works that implemented learning-based approaches such as in~\cite{bauza2019localization}, where the shape of the object was reconstructed from tactile imprints to identify and localize the object for in-hand manipulation, and in~\cite{villalonga2021tactile}, where object pose estimates were determined using geometric contact rendering. Other works include using a network trained on simulated contact shapes to obtain the pose distribution~\cite{bauza2022tac2pose} and object recognition by multi-modal associations~\cite{lin2019learning}. Tactile perception also includes studies such as active visuo-tactile point cloud registration for pose estimation between sparse point clouds computed by filter-based methods~\cite{murali2021active}, or combining vision and tactile sensors to conduct visual servoing and localization to improve the estimation accuracy~\cite{chaudhury2022using}.

Object shape estimation~\cite{de2021simultaneous, matsubara2017active} is another implementation of vision-based tactile sensors. In~\cite{de2021simultaneous}, the authors aim to plan grasps by exploration from multiple touches and also claim that an initial grasp attempt based on the initial guess of the overall object shape can provide information on the far side of the object for shape estimation, that allows probabilistic approaches to determine the next grasp location.

Pose estimation errors in current works are usually too large to be used for practical tasks when only tactile sensors are used, as the error ranges from 5mm to 60mm in~\cite{villalonga2021tactile} and the main dimension error was around 5\% for reconstructed known objects in~\cite{bauza2019localization}. In addition, a survey on robot tactile perception noted that high-accuracy localization might not have been achieved~\cite{luo2017robotic}. Another survey observed that often easier to provide data from contact-based interactions than to pre-define an accurate analytical model~\cite{li2020review}. Thus many methods tend to be data-driven and object dependent which poses certain challenges during practical implementations especially when the objects at task are constantly changing.

Two key challenges in using these sensors are: (1) information provided by a sensor is very limited due a small sensing area that cannot achieve reasonable feature matching~\cite{chaudhury2022using, luo2017robotic}, and (2) contact non-uniqueness, where a contact is ambiguous due to resemblance to other contacts from another pose of the same or different object, as illustrated in~\cite{bauza2022tac2pose, lin2019learning}.

Hence, we propose to use pattern augmentation on 3DP parts to provide a basis for object recognition and pose refinement, by leveraging the advantage of 3DP since the parts are also manufactured by 3DP. Small and unique patterns are rich in feature information to prevent contact non-uniqueness in vision-based tactile sensors. The motivation stems from manual labels on parts for distinguishing between parts, e.g. imprinting names on shoe insoles.



\section{Methodology} \label{sec:method}

This section discusses the creation of the pattern library and the overall workflow for object recognition and pose refinement of 3DP parts in practical tasks.

\subsection{Overall pipeline} \label{ssec:implementation}

Our pipeline (Fig.~\ref{fig:pipeline}) shows the estimation of an initial pose of an object by a 3D camera to conduct grasping. After grasping, a vision-based tactile sensor captures the image of the imprint and the indentation point cloud. Image segmentation is performed on the imprint image to obtain the pattern mask. An example is Segment Anything Model (SAM)~\cite{sam}, an AI model that can "cut-out" all objects in an image. The pattern mask would be used to conduct object classification and pose refinement. As each pattern in the library corresponds to an object, the original geometrical shapes of the objects would not be necessary for object identification. 

The object class label, $L$, can be obtained with the Intersection over Union (IoU) loss~\cite{iouloss} of the actual imprint $I$, against all other $j^{th}$ pattern in the pattern library ($S$) where:
\begin{equation} L = \min_{P_j \in S} (1 - \frac{(I \cap P_j)}{(I \cup P_j)})\ \end{equation}

The IoU is an evaluation metric to measure the overlap of two regions, or patterns. A smaller IoU loss value indicates better similarity of $I$ to $P_j$. Images of the patterns are also processed with a morphological transformation, and dilation, by an elliptical structuring element to mimic the smooth corners present in the actual imprint.


The actual imprint point cloud is cropped by its image mask and scaled to real-world values. Point cloud registration, such as FilterReg~\cite{gao2019filterreg}, is computed between the imprint point cloud and its corresponding point cloud in the pattern library that was identified during classification. The registration transformation (Fig.~\ref{fig:pipeline}), is the transformation of the source (identified pattern point cloud from our library) to the target (imprint point cloud). As such, pose refinement can be conducted as the transformation of the pattern w.r.t to the object can be obtained during the augmentation phase in Section~\ref{ssec:create_patterns}. To improve the accuracy and computation time of the point cloud registration, the source point cloud is also subjected to an initial transformation by translating its centroid to the centroid of a box that bounds the mask of the imprint.

\subsection{Creation and augmentation of pattern library} \label{ssec:create_patterns}

\begin{figure}[t!]
    \centering
	\includegraphics[width=0.45\textwidth]{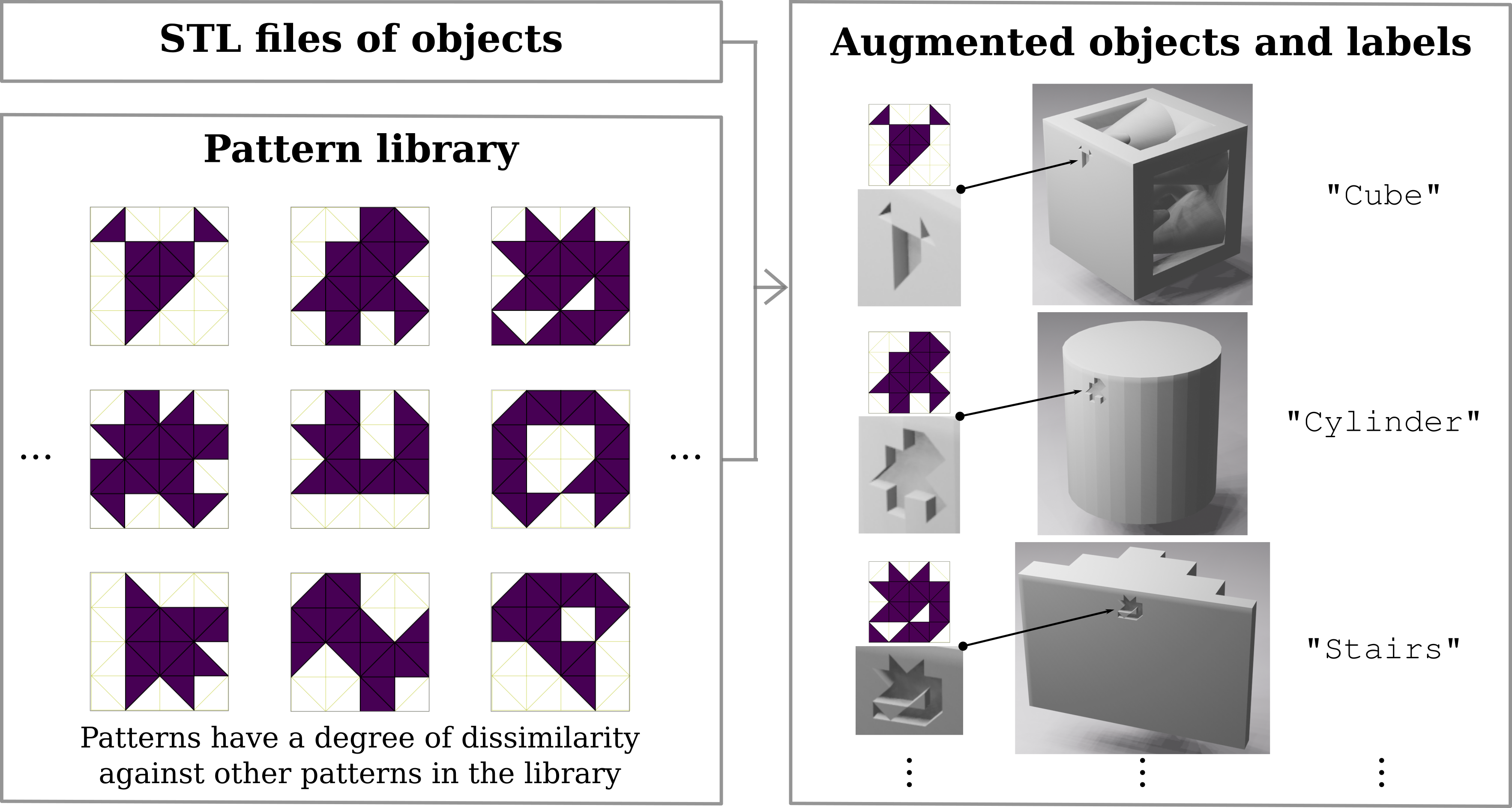}
        \caption{A unique pattern library is obtained by using simulated annealing to place triangles on a grid. The pattern library and the STL files of the objects are used to create pattern-augmented objects and their corresponding labels. Labels correspond to patterns rather than objects.}
        \label{fig:pattern_dataset}
\end{figure}

Small and unique features would aid feature matching in vision-based tactile sensors as discussed in Section~\ref{sec:related_work}. Thus, we propose the idea of augmenting small and unique patterns on 3DP parts to aid object recognition and pose estimation. In~\cite{gartus2013small, gartus2020experts}, abstract patterns were created by placing triangular elements on a rectangular grid using a simulated annealing stochastic optimization algorithm~\cite{kirkpatrick1983optimization}. We adopted the idea to optimize the triangular placements but with a Delaunay triangulation grid obtained by staggered row sampling~\cite{delaunay} instead of a rectangular grid in~\cite{gartus2013small, gartus2020experts}. 

Patterns are generated by finding the triangle placement that can meet a target connectivity, which is the optimization objective for simulated annealing, used together with a linear multiplicative cooling function. A random number of triangles, $N$, is selected for every pattern, and the target connectivity is a random number between $[N-2, N]$. Specifically, the connectivity is the number of triangles connected to their neighbors using a graph search. Higher connectivity seems desirable as empty regions between unconnected triangles may cause the formation of sub-patterns that may result in the non-uniqueness of the patterns. This might affect object recognition during feature matching. To ensure a certain degree of dissimilarity, or dispersion ($\delta$) is present between patterns of the library, we use a distance measure, $d(P_i,P_j)$, based on Hu Moments~\cite{humoment} to conduct shape matching of a new pattern sample $P_i$ against all other $P_j$ patterns in the library, where $H_{m,i}$ and $H_{m,j}$ be the $m^{th}$ log transformed Hu Moment for $P_i$ and $P_j$. A smaller distance indicates greater similarity. 
\begin{equation} d(P_i,P_j) = \sum_{m=0}^{6}\frac{|H_{m,i}-H_{m,j}|}{|H_{m,i}|}\ \end{equation}
Next, for every new $P_i$, we ensure that the minimum dispersion of the pattern library, $S$, is greater than a threshold, $\alpha$.
\begin{equation} \delta(S) = \min_{P_i, P_j \in S} {d(P_i, P_j) > \alpha} \end{equation}


Pattern augmentation can be performed on the objects to print with the pattern library. The augmentation locations of the patterns are fixed at the center of the plane on the side of the object, and offset by a small distance from the top edge, e.g. 1mm distance. These objects were properly orientated during the design phase to enable automated augmentation of the patterns. \textit{Blender} was used to conduct Boolean difference on the objects with the pattern STL files automatically, thus creating imprints of 1mm depth on the objects (Fig.~\ref{fig:pattern_dataset}). Labels are automatically created by referencing the pattern number with the name of the object STL file.

Our pattern library of 1095 patterns was created with $N=[10,20]$ on a 4x4 square Delaunay triangulation grid with $\alpha=0.1$ using the libraries \textit{Matplotlib} and \textit{Scipy}. Some examples of the patterns obtained are shown in Fig.~\ref{fig:pattern_dataset}. Note that the grid size can be changed and the number of patterns in the library can be increased, as the number of patterns selected for this library is arbitrary. Expansion of a particular library could also be performed by computing $\delta(S)$ for every new $P_i$. The pattern size can be easily changed by scaling the grid. Our pattern size was scaled to 5mm and used in all experiments. The \textit{trimesh} library was used to obtain the STL files of the patterns and subdivide the meshes to have more vertices. The value used for subdivision was 0.1. These vertices are translated into voxelized point clouds using \textit{Open3D} library, to be used in point cloud registration in the subsequent steps.

\section{Experiments} \label{sec:experiments}

We evaluate the effectiveness of pattern augmentation for 3DP parts in object recognition and pose refinement from three perspectives: (1) Evaluation of robustness of pattern augmentation technique, (2) Evaluation of insertion success rate and pose refinement accuracy, and (3) Evaluation with real insertion tasks to mimic packing parts into shadow boxes.

\subsection{Specifications}

We list some specifications used. In all experiments, a Universal Robot (UR5e) equipped with a Robotiq Hand-E parallel gripper with a flat finger and a GelSight Mini tactile sensor was used. Specifications of the workstation used are Intel Core i7-6700HQ CPU at 2.60GHz $\bigtimes$ 8 with an NVIDIA Quadro M1000M graphics card. All objects were printed using the HP MJF5200 printer with Nylon powder.

\subsection{Evaluation of pattern augmentation technique} \label{ssec:classification}

The robustness of the pattern augmentation technique was evaluated by conducting part identification for 30 randomly selected patterns from the library of 1095 patterns. Each pattern was augmented on the same cube as shown in Fig.~\ref{fig:overview}, to depict a unique part. The cubes were grasped in the same initial position to capture the imprints and classification was executed with the procedure in Fig.~\ref{fig:pipeline}. All 30 imprints were identified correctly, which illustrates the robustness of our pattern augmentation technique, where a certain degree of dissimilarity between the patterns in the library was ensured.

\subsection{Evaluation of success rate and accuracy} \label{ssec:refinement_accuracy}

\begin{table*}[!t]
\renewcommand{\arraystretch}{1}
\caption{Evaluating insertion success rate with physical experiments.}
\label{table:holes}
\centering
\begin{tabular}{|c|c|c|c|c|c|c|c|c|c|c|c|c|c|}
\hline
& \multicolumn{6}{c|}{\textbf{Insertion of 30.2mm cube into 31.6mm hole}}&\multicolumn{6}{c|}{\textbf{Insertion of 30.2mm cube into 30.7mm hole}}\\
\hline
\# & \makecell{X\\(mm)} & \makecell{Y\\(mm)} & \makecell{$\theta_z$\\($^\circ$)} & \makecell{$Y_{ref}$\\(mm)} & \makecell{Insert with\\refinement} & \makecell{Insert w/o\\refinement} & \makecell{X\\(mm)} & \makecell{Y\\(mm)} & \makecell{$\theta_z$\\($^\circ$)} & \makecell{$Y_{ref}$\\(mm)} & \makecell{Insert with\\refinement} & \makecell{Insert w/o\\refinement}\\
\hline
1 & -1.713 &  -1.747 & -2.445 & -3.396 &  \checkmark & $\bigtimes$  & -0.881 & -0.739 & -2.965 & -1.367 &  \checkmark & $\bigtimes$\\
\hline
2 & -0.046 & -0.990 & -1.615 & -0.691 &  \checkmark & $\bigtimes$   & 1.671 & 1.950 & -1.601 & 1.399 &  \checkmark & $\bigtimes$\\
\hline
3 & -0.514 & -0.492 & -0.803 & -1.756 &  \checkmark & $\bigtimes$   & -1.485 & 1.061 & -1.578 & 1.280 &  $\bigtimes$ & $\bigtimes$\\
\hline
4 & -0.567 & -0.755 & -1.187 & -0.892 &  \checkmark & \checkmark    & -0.466 & -2.488 & 1.161 & -2.695 &  \checkmark & $\bigtimes$\\
\hline
5 & 1.195 & 1.360 & -2.487 & 0.132 &  $\bigtimes$ & $\bigtimes$     & -0.625 & 0.627 & 2.912 & 0.740 &  \checkmark & \checkmark\\
\hline
6 & 2.406 & 0.817 & 2.648 & 0.361 &  \checkmark & \checkmark        & -0.634 & 1.959 & -1.897 & 1.100 &  $\bigtimes$ & $\bigtimes$\\
\hline
7 & 1.799 & 1.242 & -0.701 & 0.557 &  \checkmark & $\bigtimes$      & -1.021 & -2.229 & -0.525 & -2.514 &  \checkmark & $\bigtimes$\\
\hline
8 & -0.625 & -2.182 & -2.710 & -3.235 &  \checkmark & $\bigtimes$   & -0.328 & 2.081 & -0.186 & 1.335 &  \checkmark & $\bigtimes$\\
\hline
9 & -1.999 & 2.129 & -2.813 & 0.746 &  \checkmark & $\bigtimes$     & 0.134 & -1.210 & -2.487 & -1.950 &  $\bigtimes$ & $\bigtimes$\\
\hline
10 & -1.457 & 1.693 & 1.274 & 1.360 &  \checkmark & $\bigtimes$     & -1.669 & -2.251 & 1.749 & -2.384 &  \checkmark & $\bigtimes$\\
\hline
11 & -1.589 & 1.567 & 2.026 & 1.668 &  \checkmark & \checkmark      & -0.690 & 1.869 & 0.792 & 0.549 &  \checkmark & $\bigtimes$\\
\hline
12 & 1.281 & -2.031 & 2.218 & -1.923 &  \checkmark & $\bigtimes$    & -2.328 & 1.537 & -0.750 & 0.009 &  $\bigtimes$ & $\bigtimes$\\
\hline
13 & 1.914 & -1.944 & -0.511 & -2.468 &  \checkmark & $\bigtimes$   & 0.481 & 1.714 & 0.642 & 1.384 &  \checkmark & $\bigtimes$\\
\hline
14 & -0.075 & -0.093 & 0.269 & -0.713 &  \checkmark & \checkmark    & 2.222 & -1.305 & 0.660 & -1.721 &  \checkmark & $\bigtimes$\\
\hline
15 & 2.337 & -1.786 & 1.989 & -1.925 &  \checkmark & $\bigtimes$    & -0.670 & 2.310 & 2.552 & 2.888 &  $\bigtimes$ & $\bigtimes$\\
\hline
16 & 2.186 & -1.585 & -0.208 & -0.972 &  \checkmark & $\bigtimes$   & 2.067 & 2.278 & -2.948 & 1.904 &  $\bigtimes$ & $\bigtimes$\\
\hline
17 & 1.124 & -1.970 & 1.853 & -2.117 &  \checkmark & $\bigtimes$    & -1.004 & 0.758 & -2.152 & -0.504 &  $\bigtimes$ & $\bigtimes$\\
\hline
18 & 1.129 & -2.493 & -2.586 & -2.993 &  \checkmark & $\bigtimes$   & 0.752 & -2.479 & 1.140 & -2.707 &  \checkmark & $\bigtimes$\\
\hline
19 & -1.708 & -1.829 & 2.113 & -2.065 &  \checkmark & $\bigtimes$   & 1.355 & 1.094 & 1.606 & 1.451 &  \checkmark & $\bigtimes$\\
\hline
20 & 1.041 & -0.008 & 1.2563 & -0.429 &  \checkmark & \checkmark    & 2.149 & 0.0980 & -1.706 & -0.067 &  \checkmark & \checkmark\\
\hline
& \multicolumn{6}{c|}{\textbf{Success rate: From 25\% to 95\% with refinement}}&\multicolumn{6}{c|}{\textbf{Success rate: From 10\% to 60\% with refinement}}\\
\hline
\end{tabular}
\end{table*}

\begin{figure}[ht] 
\centering\begin{tabular}{c}
\subfloat[]{\includegraphics[height=2.7cm]{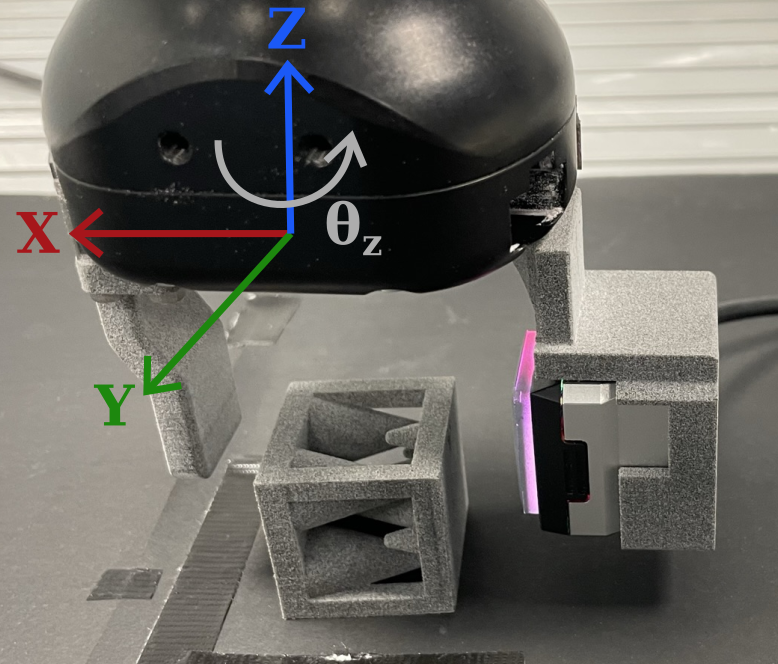}}\hspace{.2cm}
\subfloat[]{\includegraphics[height=2.7cm]{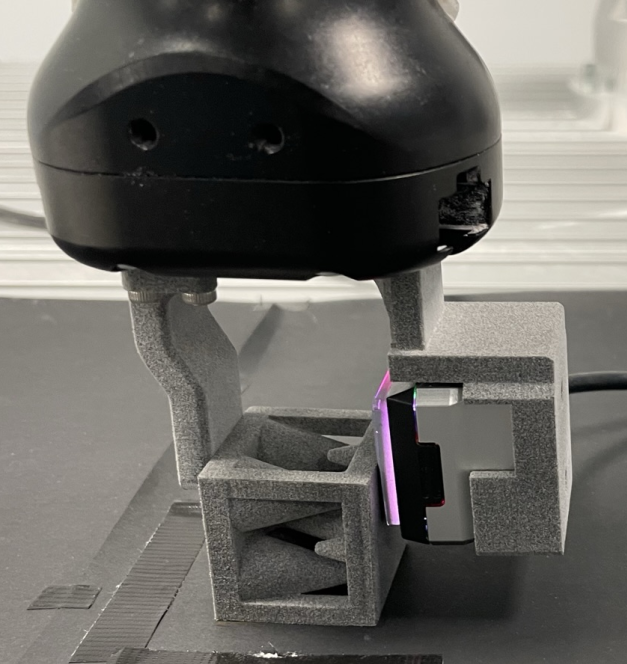}}%
\end{tabular}%
\caption{Random initial pose of robot manipulator: (a) Illustration of perturbation axes; (b) Cube initial position is unknown after grasping which resulted from the random perturbation of robot manipulator.} 
\label{fig:grasping}
\end{figure}

The evaluation of insertion success rate and pose refinement accuracy was conducted with a physical peg-in-hole insertion task. The objective was to measure the insertion success rate when the robot manipulator was subjected to random perturbations. Specifications of the experiment are listed below:

\begin{itemize}
    \item Insertion peg was a square cube measuring 30.2mm.
    \item Dimensions of square holes were 31.6mm and 30.7mm.
    \item Initial pose of gripper was subjected to random perturbations of (X, Y, $\theta_z$), where X and Y ranges between [-2.5mm, 2.5mm] and $\theta_z$ ranges from [-3$^\circ$, 3$^\circ$] (Fig.~\ref{fig:grasping}a).
\end{itemize}

The initial position of the cube is unknown after grasping due to the random perturbation of the robot manipulator (Fig.~\ref{fig:grasping}b). However, the position of the cube relative to the gripper can be extracted from the vision-based tactile sensor by point cloud registration between the real pattern imprint and the voxelized point cloud from the pattern library, thus allowing pose refinement for successful insertion which was discussed in Section~\ref{ssec:implementation}. In a typical insertion task by picking an object from a plane, the pose refinement needed is the translation on the X-axis, Y-axis, and rotation $\theta_z$. During grasping, the gripper fingers push the object to its centroid thus the offset of the object's centroid on the X-axis would be zero. In addition, $\theta_z$ could be obtained by extracting the rotation of the gripper. Hence, the only unknown variable needed is translation on the Y-axis, namely the Y-refinement ($Y_{ref}$).

The robot attempted 20 insertions for each hole dimension and the results are in Table~\ref{table:holes}, which illustrates a large improvement in success rate with pose refinement, which may be attributed to the unique features of the patterns that can be well-captured by the vision-based tactile sensor. From the insertion experiment for the 31.6mm hole in Table~\ref{table:holes}, it can be seen that the refinement magnitude can be rather large at $>$3mm, while the hole allowance was only 1.4mm which indicates the effectiveness of our pattern augmentation method. Additionally, we were able to achieve a  high success rate of 95\% for a tight hole allowance of 1.4mm. Note that the $Y_{ref}$ does not equate to the random Y perturbation of the manipulator as the actual $Y_{ref}$ needed by the object would be affected by the rotation of gripper ($\theta_z$), as the gripper fingers will push the object during grasping. 

To measure the pose refinement accuracy, we did experiments where only a Y-offset was applied to the manipulator. The target refinement value is the offset and the resulting $Y_{ref}$ is shown in Table~\ref{table:accuracy}, indicating good accuracy due to low percentage errors in sub-millimeter ranges. Thus, our method is able to conduct pose refinement of sub-millimeter accuracy.

\begin{table}[ht]
    \renewcommand{\arraystretch}{1}
    \caption{Evaluating pose refinement accuracy.}
    \centering
    \begin{tabular}{|c|c|c|c|c|c|c|}
    \hline
    Y-offset (mm)  & 1.0  & 2.0 & 3.0 & -1.0 & -2.0 & -3.0 \\
    \hline
    $Y_{ref}$ (mm) & 0.846 & 1.715 & 2.605 & -0.976 & -2.060 & -2.964\\
    \hline
    Error (\%) & 15.4 & 14.3 & 13.2 & 2.4 & 3.0 & 1.2 \\
    \hline
    \end{tabular}
    \label{table:accuracy}
\end{table}


\subsection{Evaluation of the implementation for robotic tasks}

\begin{figure}%
\centering\begin{tabular}{c}{\includegraphics[width=0.40\textwidth]{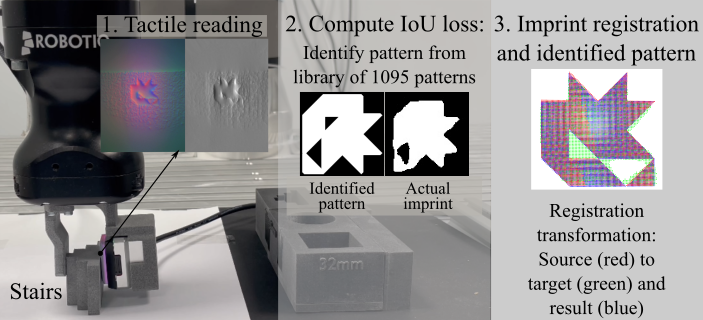}}\\
\end{tabular}
\caption{Robotic sorting and packing into shadow boxes. Three objects were shown in the video (\protect\url{https://youtu.be/3e6gvkZUk8c}) and the dimensional allowance between the objects and holes ranges from 1.3mm to 2.3mm.}
\label{fig:shademo}%
\end{figure}

The evaluation of the pattern augmentation method was conducted by physical insertion tasks that mimics robotic sorting and packing (Fig.~\ref{fig:shademo}). Specifically, three 3DP parts with augmented patterns (Fig.~\ref{fig:pattern_dataset}) were placed at a random position on a table and the robot picked and packed them in their respective shadow boxes or holes. Their dimensions (in mm) are below. Note that the stairs and cube were real samples from HP Labs used for certain industrial tasks, and the HP MJF5200 printer has sub-millimeter tolerances.

\begin{enumerate}
    \item Stairs with L46.2 by W20.3 into L48.5 by W22.3 hole.
    \item Cylinder with $\diameter$30.2 into $\diameter$31.5 hole.
    \item Cube with L30.2 into L31.6 hole.
\end{enumerate}

In the experiment, an initial pose estimate of the object was obtained by an L515 Intel RealSense depth camera for the robot to conduct grasping. Upon grasping, the vision-based tactile sensor provides the RGB image and point cloud of the pattern imprint. As discussed in Section~\ref{ssec:implementation}, the pattern imprint would be matched with the pattern library to get the correct object class label and the refinement transformation required, which only took 0.4s once the pattern mask was obtained. In addition, although only three objects were used in the experiment, each pattern was matched to a pattern library of 1095 patterns and was still able to quickly identify the correct labels. Note that these patterns used were different from the 30 patterns used in Section~\ref{ssec:classification}. Due to the set-up of the experiment, we would only need to compensate along the Y-axis as mentioned in Section~\ref{ssec:refinement_accuracy}. The robot then moves to the correct shadow box, conducts pose refinement, and successfully inserts all objects into their respective shadow boxes. Thus, this practical example shows that pattern augmentation on 3DP parts is a viable method to achieve grasping, part identification, and pose refinement in one-shot robotic tasks.

We used SAM~\cite{sam} to obtain the pattern mask. Although SAM is non-specific and claimed to be unachievable in real-time, real-time performance could be achieved with specific models like Mask R-CNN~\cite{he2018mask} which could return in 0.2s, or using the improved model, Fast SAM~\cite{zhao2023fast}, that claims to be 50 times faster than SAM. In total, our approach should take less than 0.6s, which is faster than any approach that relies on a middle station for precise vision-based pose estimation.

To further illustrate the advantages of our method, we did analytical comparisons with similar works discussed in Section~\ref{sec:related_work}. We compare the pose estimation accuracy with vision-based tactile sensors, where these sensors extract the poses from tactile images of the geometries of the objects. The minimum error in~\cite{villalonga2021tactile} was 5mm, and the error was 5\% in~\cite{bauza2019localization}, which translates to 1.52mm for our 30.2mm cube. These methods are also object-dependent which increases the difficulty to transfer to an industrial domain. Comparatively, our method is not object-dependent with higher accuracy.


\section{Conclusion}

Competitive additive manufacturing technologies come with a major bottleneck of manual 3DP post-processing. The ability to customize also creates unique challenges for the existing paradigm of robotics applications, thus creating limitations for end-to-end 3DP post-processing automation. Thus, we explore the use of pattern augmentation on 3DP objects to execute grasping, part identification, and pose refinement in one shot with a tactile gripper.  We experimentally evaluate our method from three perspectives, including real tasks that mimic robotic sorting and packing, and achieved excellent classification results, a high insertion success rate of 95\%, and sub-millimeter pose refinement accuracy.

At the current state, our work is limited to being implemented as part of an end-to-end manufacturing automation line rather than a standalone process, as we assume that prior information on the initial positions of the patterns was obtained upstream, eg. from prior cleaning or quality inspection processes. Hence, a possible improvement to make our work a standalone process would be to research a method to incorporate simultaneous grasp planning so that initial positions of the patterns can be quickly extracted, allowing large objects to be grasped properly. This could be another challenge for the existing production paradigm of robotics. Our current implementation also assumed that with prior grasping information, patterns would be fully captured by the sensor. To be a standalone process, there may be occlusions on the pattern imprint. Thus, further improvements to our pattern library could include optimizing the dispersion of patterns or enhancing the pattern generation, so that pattern subsets would not correspond to another pattern in the library to reduce erroneous identification in standalone processes.

\section*{Acknowledgments}

This research was conducted in collaboration with HP Inc. and supported by National Research Foundation (NRF) Singapore through the Industry Alignment Grant (I1801E0028).

\bibliographystyle{IEEEtran}
\bibliography{IEEEabrv, ref}

\end{document}